\documentclass{article}
\usepackage{ijcai18}

\usepackage{times}
\usepackage{xcolor}
\usepackage{soul}
\usepackage[utf8]{inputenc}
\usepackage[small]{caption}
\usepackage{amsmath}
\usepackage{amsfonts}
\usepackage{amsthm}
\usepackage{amssymb}
\usepackage[english]{babel}
\usepackage{algorithm}
\usepackage{algpseudocode}
\usepackage{multirow}
\usepackage{graphics} 
\usepackage{epsfig} 

\usepackage{graphicx}
\usepackage{caption}
\usepackage{subcaption}

\title{DiGrad: Multi-Task Reinforcement Learning with Shared Actions%
}

\author{
Parijat Dewangan$^1$\thanks{These authors contributed equally.}, Phaniteja S$^1$\footnotemark[1],  
K Madhava Krishna$^1$, Abhishek Sarkar$^1$, \\
\textbf{Balaraman Ravindran$^2$} \\
$^1$ Robotics Research Center, International Institute of Information Technology, Hyderabad \\
$^2$ Indian Institute of Technology, Madras\\
\{ parijat10, phaniteja.sp\}@gmail.com,
\{ mkrishna, abhishek.sarkar\}@iiit.ac.in,\\
ravi@cse.iitm.ac.in
}

\begin{document}

\maketitle

\begin{abstract}
Most reinforcement learning algorithms are inefficient for learning multiple tasks in complex robotic systems, where different tasks share a set of actions. In such environments a compound policy may be learnt with shared neural network parameters, which performs multiple tasks concurrently. However such compound policy may get biased towards a task or the gradients from different tasks negate each other, making the learning unstable and sometimes less data efficient. In this paper, we propose a new approach for simultaneous training of multiple tasks sharing a set of common actions in continuous action spaces, which we call as DiGrad (\textbf{Di}fferential Policy \textbf{Grad}ient). 
The proposed framework is based on differential policy gradients and can accommodate multi-task learning in a single actor-critic network.
We also propose a simple heuristic in the differential policy gradient update to further improve the learning. The proposed architecture was tested on 8 link planar manipulator and 27 degrees of freedom(DoF) Humanoid for learning multi-goal reachability tasks for 3 and 2 end effectors respectively. We show that our approach supports efficient multi-task learning in complex robotic systems, outperforming related methods in continuous action spaces. 

\end{abstract}
\section{Introduction}
There has been an increasing demand for reinforcement learning (RL)\cite{sutton} in the fields of robotics and intelligent systems. Reinforcement learning deals with learning actions in a given environment to achieve a goal. Classic reinforcement learning techniques make use of linear approximation or tabular methods to learn such correlation\cite{konidaris2011value}.

With the advancements of deep neural networks in the recent times, learning non-linear approximations and feature extraction has becomes much simpler. It was believed that non-linear approximators like neural network are hard to train in reinforcement learning scenario. However recent advancements in RL has successfully combined the deep neural networks with RL and stabilized the learning process. Deep Q Networks(DQN) \cite{dqn} used Convolutional neural networks(CNN) and fully connected layers to make the RL agents learn to play the ATARI games. Following the success of DQN, several improvements on this architecture like Double DQN\cite{ddqn}, Prioritized Replay\cite{per}, Duelling Network\cite{dueling} are proposed which propelled the use of Deep RL in multi-agents. \cite{lillicrap2015continuous} proposed Deep Deterministic Policy Gradient(DDPG) for continuous control tasks which further extended the scope of Deep RL applications to robotics.
\begin{figure}\label{framework}
\includegraphics[width=\columnwidth]{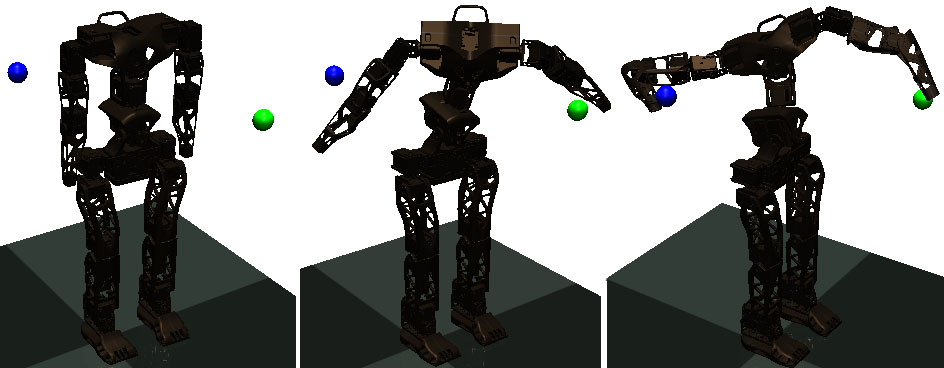}
\caption{Humanoid robot multi-tasking to reach two goals simultaneously. The two goals are shown using blue and green coloured balls.}
\end{figure}

Robotic systems like open and closed kinematic chains offer fresh perspectives to employ deep RL algorithms. \cite{door} applied DDPG framework to learn manipulation tasks in complex 3D environments. Following this, \cite{humanoid} applied DDPG to learn reachability tasks in Humanoid robot. Even though they are not multi-agent systems, they can be posed as multi-tasking systems where there are shared actions (common kinematic chains). As shown in Fig. 1, the spine/torso is the common chain which contributes to the reachability tasks of both the hands in humanoid robot. In this paper, we propose a novel framework called DiGrad based on differential policy gradients to learn multi-tasking in such robotic systems where different tasks may share a set of common actions. 

There are several mathematical approaches which try to address the problem of multi-tasking in branched manipulators. However, classical methods based on Jacobian\cite{jacob,bfgs} has very limited capability in branched manipulators. Methods like Augmented Jacobian\cite{siciliano1990kinematic,chiaverini1997singularity} have constrained solution spaces, while methods based on optimization\cite{klein1995new} often doesn't provide real time control. Hence, RL based solvers are of great use in this domain which can sample the entire solution space and learn a real time controller in such complex robotic systems.

One direction for learning multiple tasks in such scenarios is to use DDPG to learn a compound policy taking care of all the tasks. However we found DDPG to be unstable for such multi-task scenarios. DiGrad addresses this problems by using differential policy gradient updates. We test our framework on branched manipulators shown in Fig. \ref{8link} for learning reachability tasks of multiple end effectors simultaneously. The proposed framework shows substantial improvement over DDPG and is considerably robust on all the experiments we conducted.

The rest of the paper is organised as follows. Section 2 and 3 discusses the related works and background. Section 4 explains the mathematical background behind the proposed framework and provides the detailed algorithm. Finally, Section 5 and 6 contain the experimental results and discussion respectively.

\section{Related Works}
Most of the multi-task reinforcement learning algorithms rely on transfer learning approaches. \cite{lazaric2012transfer} shows a good collection of these methods. Some of the recent works based on this approach are \cite{rusu2015policy,yin2017knowledge}. Some works by \cite{borsa2016learning} and \cite{zhang2016deep} explored learning universal abstractions of state-action pairs or feature successors. 

Apart from transfer learning, some works like \cite{lazaric2010bayesian}, \cite{dimitrakakis2011bayesian} investigated joint training of multiple value functions or policies. In a deep neural network setting, \cite{teh2017distral} provided a framework for simultaneous training of multiple stochastic policies and a distilled master policy. Unlike our work, \cite{teh2017distral} uses multiple networks for each policy and one more network for the distilled policy. In our work, we show how we can use a single network to learn multiple deterministic policies simultaneously.

All the above mentioned methods assume multi-agent scenario whereas in our paper, we concentrate on learning multiple tasks in a robotic system. Some very recent works in this scenario are  \cite{yang2017multi} and \cite{kulkarni2016hierarchical}. These works do not talk about the actions which are shared among different tasks, thus limiting their applicability. Unlike these frameworks, we explore the case of multi-task learning in branched manipulator which have shared action-spaces.
\section{Background}
We consider a standard reinforcement learning setup consisting of an agent interacting with an environment $E$ in discrete time steps. At each time step $t$, the agent takes a state $s_t \in S$ as the input, performs an action $a_t \in A$ according to a policy, $\mu: S \to A$ and receives a reward $r_t \in R$. We assume a fully observable environment and model it as a Markov decision process with state space $S$, action space $A = {\rm \textit{I}\!\textit{R}}^N$, an initial state distribution $p(s_1)$, state transition dynamics $p(s_{t+1}|s_t,a_t)$ and a reward function $r(s_t,a_t)$.

The goal in reinforcement learning is to learn a policy $\mu$ which maximizes the expected return $R_t = \sum\limits_{t>0}\gamma^tr(s_t,a_t)$, where $\gamma \in [0,1]$ is the discount factor. Since the return depends on the action taken, the policy may be stochastic but in our work we consider only deterministic policies. Hence the discounted state visitation distribution for a policy $\mu$ is denoted as $\rho^\mu$.

The action-value function used in many reinforcement learning algorithms is described as the expected return after taking an action $a_t$ in state $s_t$ and thereafter following the given policy:
$$Q^\mu(s_t,a_t) = \mathbb{E}[R_t|s_t,a_t]$$
DDPG is an off policy learning algorithm that uses an actor-critic based framework for continuous control tasks.
In DDPG, both actor and critic are approximated using neural networks ($\theta^\mu$, $\theta^Q$). The problem of instability in training is addressed by using target networks ($\theta^{\mu'}$, $\theta^{Q'}$) and experience replay using a replay buffer. In this setting, the critic network weights are optimized by minimizing the following loss:
\begin{equation}\label{loss_ddpg}
L(\theta^{Q}) = (Q(s_t,a_{t}|\theta^{Q}) - y_{t})^2
\end{equation}
where,
\begin{equation}\label{target_ddpg}
y_{t} = r(s_t, a_t) + \gamma Q'(s_{t+1},{\mu'}_{t+1}(s_{t+1}|\theta^{\mu'})|\theta^{Q'}) 
\end{equation}
The gradient ascent update on policy network (actor) is given using Deterministic Policy Gradient(DPG) theorem(\cite{silver2014deterministic}). Suppose the learning rate is $\eta$, then:
\begin{equation}
\theta^\mu = \theta^\mu + \eta\nabla_{a}Q(s_t,a_t|\theta^Q)|_{a=\mu(s_t|\theta^{\mu})}\nabla_{\theta^\mu}\mu(s|\theta^{\mu})
\end{equation}
Finally the updates on target networks are: 
\begin{equation*}
\begin{split}
\theta^{Q'} \leftarrow \tau \theta^Q + (1-\tau)\theta^{Q'} \\\theta^{\mu'} \leftarrow \tau \theta^\mu + (1-\tau)\theta^{\mu'}
\end{split}
\end{equation*}
where, $\tau << 1$.

In the proposed framework, we use some of the basic concepts of DDPG and use DPG theorem to derive a policy gradient update which can support robust multi-task learning. Finally we explain how the learning can be improved by using a simple heuristic in case of shared actions.

\section{DiGrad: Differential Policy Gradients}
\begin{figure}[h]
\center
\includegraphics[width=0.8\columnwidth]{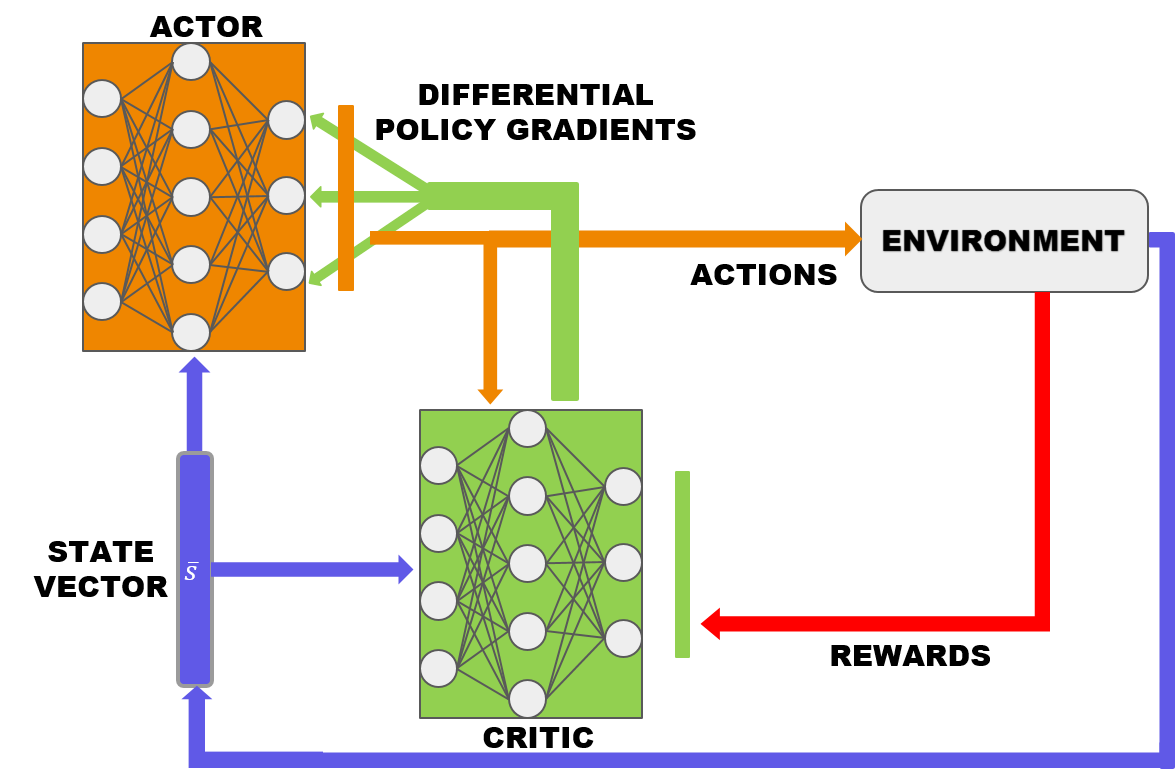}
\caption{Overview of the algorithm showing differential policy gradient update}
\label{digrad}
\end{figure}

We propose a framework for simultaneous reinforcement learning of multiple tasks shared actions in continuous domain. The method is based on differential action-value updates in actor-critic based framework using DPG theorem.
The framework learns a compound policy which optimally performs all the shared tasks simultaneously.
Fig. \ref{digrad} shows the higher level description of the method.
In this section, we describe the mathematical framework behind this work.
\subsection{Environment setting}
Consider $n$ tasks in a standard RL setting and their corresponding $n$ action spaces $A_1$, $A_2$, ..., $A_n$. We will assume that the state space $S$ is the same across all the tasks and an infinite horizon with discount factor $\gamma$. Let $A$ be a compound action space which performs the given set of $n$ tasks simultaneously. Let $a \in A$ denote compound actions, $a_i \in A_i$ denote actions and $s \in S$ denote states. Therefore the relation between the compound actions and all the $n$ actions is given by,
$$a = \bigcup\limits_{i=1}^{n}a_i.$$
Suppose $a_s$ corresponds to the set of actions which affects $k$ of the $n$ tasks, then $a_s$ is given by
\begin{equation}\label{share_action}
\begin{split}
a_s = \bigcap\limits_{j=1}^k a_j \qquad \textnormal{where,} \quad a_j \subset a
\end{split}
\end{equation}
The reward functions for the $i^{th}$ task depend only on the corresponding actions $a_i$. Therefore, we denote the reward function as $r_i(s,a_i)$ for each task $i$; let $Q_i(s, a_i)$ be the corresponding action value function. Let $\mu$ be the compound deterministic policy and $\mu_i$ be the task-specific deterministic policies; therefore $\mu_i \subset \mu$, where 
\begin{equation*}
\begin{split}
\mu_i(s) = a_i \quad \textnormal{and} \quad \mu(s) = a
\end{split}
\end{equation*}

\subsection{Proposed framework} 
An actor-critic based framework is proposed for multi-task learning in the given environment setting. We use a compound policy $\mu(s)$ parametrized by $\theta^\mu$, instead of multiple policy networks for each policy $\mu_i$. 
$$\mu:S \rightarrow A$$
A simple parametrization for action-value functions would be to have a separate network $\theta^{Q_i}$ for each $Q_i(s,a_i)$, which outputs an action-value corresponding to the $i^{th}$ task.
Another approach for modelling action-value function is to have a single network parametrized by $\theta^Q$ which outputs action-values for all the tasks. Here, $Q_i$ is the action value corresponding to the $i^{th}$ task. 
\cite{yang2017multi} showed that a single critic network is sufficient in multi-policy learning. In this setting, the penultimate layer weights for each $Q_i$ are updated based on only the reward $r_i(s,a_i)$ obtained by performing the corresponding action $a_i$. The remaining shared network captures the correlation between the different actions $a_i$. Hence this kind of parametrization is more useful in the case of shared actions. Apart from this, the number of parameters are significantly reduced by the use of a single network. 
\subsection{Critic update}
Consider a single actor-critic based framework where critic $Q(s,a)$ is given by function approximators parametrized by $\theta^Q$ and the actor $\mu(s)$ is parametrized by $\theta^\mu$. Let the corresponding target networks be parametrized by $\theta^{\mu'}$, $\theta^{Q'}$. Since we have multiple critic outputs $Q_i$, we optimize $\theta^Q$ by minimizing the loss as given by \cite{yang2017multi}:
\begin{equation}\label{loss}
L(\theta^{Q}) = \sum_{i=1}^{n} (Q_i(s_t,a_{it}|\theta^{Q}) - y_{it})^2
\end{equation}
where the target $y_{it}$ is
\begin{equation}\label{target}
y_{it} = r_i(s_t, a_t) + \gamma Q'(s_{t+1},{\mu'}_{t+1}(s_{t+1}|\theta^{\mu'})|\theta^{Q'}) 
\end{equation}
If there are multiple critic networks, network parameters are optimized by minimizing the corresponding loss:
\begin{equation}\label{loss1}
L(\theta^{Q_i}) = (Q_i(s_t,a_{it}|\theta^{Q}) - y_{it})^2
\end{equation}
In both the settings of critic, there is only a single actor which learns the compound policy $\mu$. The differential policy gradient update on the compound policy is explained in the next subsection.  
\subsection{Differential Policy Gradient}
Each task has a corresponding reward, $r_i(s,\mu_i(s))$ and hence to learn all the tasks we need to maximize the expected reward for each of the task with respect to the corresponding action, $a_i$. Therefore the performance objective to be maximized is: 
\begin{equation}
\label{obj_func}
J(\mu,\{\mu_i\}_{i=1}^n) = \sum_{i=1}^{n}\mathbb{E}_{s\sim \rho^\beta}[r_i(s,\mu_i(s))]
\end{equation}
The update on the parametrized compound policy $\mu(s|\theta^\mu)$ is given by applying deterministic policy gradient theorem on Eq. \eqref{obj_func}. The resulting update is:
$$
\hspace{-1.5cm}\nabla_{\theta^\mu} J \approx \sum_{i=1}^{n}\mathbb{E}_{s\sim \rho^\beta}[ \nabla_{\theta^\mu}Q_i(s,a_i|\theta^Q)|_{a_i=\mu_i(s|\theta^{\mu})}] 
$$
\begin{equation}
\label{policy_gradient}
\begin{split}
\hspace{-0.1cm} = \sum_{i=1}^{n}\mathbb{E}_{s\sim \rho^\beta}[ \nabla_{a_i}Q_i(s,a_i|\theta^Q)|_{a_i=\mu_i(s|\theta^{\mu})}\nabla_{\theta^\mu}\mu_i(s|\theta^{\mu})]
\end{split}
\end{equation}
\subsubsection{DiGrad with shared tasks}
In the above environment setting, let all the $n$ tasks share a common set of actions $a_s$, i.e., $k=n$. Let $a_1$, $a_2$, ..., $a_k$ be the actions corresponding to these $k$-tasks. Therefore from Eq. \eqref{share_action}, 
$$a_s = \bigcap\limits_{j=1}^{k}a_j$$
Now the compound action $a$ can be written as:
\begin{equation*}
\begin{split}
\hspace{-3.6cm} a = \{a_1 \cup a_2 ...\cup a_k\}
\end{split}
\end{equation*}
\begin{equation*}
\begin{split}
= \{ \{a_1 \setminus a_s\} \cup\{ a_2 \setminus a_s\}...\cup\{ a_k \setminus a_s\}\cup a_s \}
\end{split}
\end{equation*}
\begin{equation*}
\begin{split}
\hspace{-3cm}= \{ a_1^d \cup a_2^d...\cup a_k^d\cup a_s\} 
\end{split}
\end{equation*}
where,  $a_j^d = \{ a_j \setminus a_s \}$

Here we can see that $a_1^d, a_2^d, a_k^d, a_s$ are disjoint sets. Therefore we can write $a_i$ = [$a_i^d$,$a_s$]. Similarly, $\mu_i = [\mu_i^d,\mu_s]$. From here onwards, to make it succinct we drop ${s\sim \rho^\beta}$ from the subscript of $\mathbb{E}_{s\sim \rho^\beta} $ and simply represent it as  $\mathbb{E}$.\\ \\
Substituting these in Eq. \eqref{policy_gradient}:
\begin{equation*}
\begin{split}
\nabla_{\theta^\mu} J \approx \sum_{i=1}^{k}\mathbb{E}[ \nabla_{[a_i^d, a_s]}Q_i(s,a_i|\theta^Q)|_{a_i^d=\mu_i^d(s|\theta^{\mu}), a_s=\mu_s(s|\theta^{\mu})} \\\nabla_{\theta^\mu}[\mu_i^d(s|\theta^{\mu}), \mu_s(s|\theta^{\mu})]] 
\end{split}
\end{equation*}
On expanding with respect to gradient operator we get,
\begin{equation*}
\begin{split}
= \sum_{i=1}^{k}\mathbb{E}
\begin{bmatrix}
\nabla_{a_i^d}Q_i(s,a_i|\theta^Q)|_{a_i^d=\mu_i^d(s|\theta^{\mu})}\\
\nabla_{a_s}Q_i(s,a_i|\theta^Q)|_{a_s=\mu_s(s|\theta^{\mu})}
\end{bmatrix}^T
\begin{bmatrix}
           \nabla_{\theta^\mu}\mu_i^d(s|\theta^\mu) \\
           \nabla_{\theta^\mu}\mu_s(s|\theta^\mu) 
         \end{bmatrix} 
\end{split}
\end{equation*}
\begin{equation*}
\begin{split}
\hspace{-1.2cm}
= \sum_{i=1}^{k}\mathbb{E}
[\nabla_{a_i^d}Q_i(s,a_i|\theta^Q)|_{a_i^d=\mu_i^d(s|\theta^{\mu})} \nabla_{\theta^\mu}\mu_i^d(s|\theta^\mu) \\ + \nabla_{a_s}Q_i(s,a_i|\theta^Q)|_{a_s=\mu_s(s|\theta^{\mu})}\nabla_{\theta^\mu}\mu_s(s|\theta^\mu)]
\end{split}
\end{equation*}
\begin{equation}\label{gradient_final}
\begin{split} 
\hspace{-1.5cm}
= \sum_{i=1}^{k}\mathbb{E}
[\nabla_{a_i^d}Q_i(s,a_i|\theta^Q)|_{a_i^d=\mu_i^d(s|\theta^{\mu})} \nabla_{\theta^\mu}\mu_i^d(s|\theta^\mu)] \\ + 
\mathbb{E}[\sum_{i=1}^{k}\nabla_{a_s}Q_i(s,a_i|\theta^Q)|_{a_s=\mu_s(s|\theta^{\mu})}\nabla_{\theta^\mu}\mu_s(s|\theta^\mu)]
\end{split}
\end{equation}
We can see that the second term on R.H.S of Eq.\eqref{gradient_final} will be zero if all the action spaces are disjoint, that is, $a_s = \emptyset$. Hence this framework can be used even when there are no shared actions. Since the update on the actor are the sum of gradients of different action values, we call this a differential gradient update. It is different from the standard gradient update where an actor is updated based on a single action value \cite{lillicrap2015continuous}.
\subsubsection{Heuristic of direction}
From Eq.\eqref{gradient_final}, we can see that the policy gradient update for the policy($\mu_s$) of shared action $a_s$ is taken as sum of the gradients of action values corresponding to the tasks it affects, whereas for policy($\mu_j^d$) of other actions $a^d_j$, the gradient update is taken as the gradient of only the corresponding $Q_j$. Thus, this uneven scaling of gradients may lead to delayed convergence and sometimes biasing. In order to equally scale all the gradient updates, we take the average value of the gradients obtained from the different Q-values for the shared action $a_s$. This modification will not affect the direction of gradient, only the magnitude will be scaled. By applying this heuristic, the differential gradient update on shared actions becomes:
\begin{equation}\label{hu}
\hspace{-0.5cm} 
\begin{split} 
\nabla_{\theta^\mu} J \approx  \sum_{i=1}^{k}\mathbb{E}
[\nabla_{a_i^d}Q_i(s,a_i|\theta^Q)|_{a_i^d=\mu_i^d(s|\theta^{\mu})} \nabla_{\theta^\mu}\mu_i^d(s|\theta^\mu)] \\ + 
\frac{1}{k}\mathbb{E}[\sum_{i=1}^{k}\nabla_{a_s}Q_i(s,a_i|\theta^Q)|_{a_s=\mu_s(s|\theta^{\mu})}\nabla_{\theta^\mu}\mu_s(s|\theta^\mu)]
\end{split}
\end{equation}
\subsubsection{Generalisation}
Suppose there are $k$ tasks which share a set of action $a_s$ as above and $n-k$ tasks which are independent, with corresponding actions $a_{k+1},... ,a_n$, then Eq. \eqref{hu} can be written as:
\begin{equation}\label{actor_update}
\hspace{-0.5cm} 
\begin{split} 
\nabla_{\theta^\mu} J \approx  \sum_{i=1}^{k}\mathbb{E}
[\nabla_{a_i^d}Q_i(s,a_i|\theta^Q)|_{a_i^d=\mu_i^d(s|\theta^{\mu})} \nabla_{\theta^\mu}\mu_i^d(s|\theta^\mu)] \\ + 
\frac{1}{k}\mathbb{E}[\sum_{i=1}^{k}\nabla_{a_s}Q_i(s,a_i|\theta^Q)|_{a_s=\mu_s(s|\theta^{\mu})}\nabla_{\theta^\mu}\mu_s(s|\theta^\mu)]\\ + \sum_{i=k+1}^{n}\mathbb{E}[ \nabla_{a_i}Q_i(s,a_i|\theta^Q)|_{a_i=\mu_i(s|\theta^{\mu})}\nabla_{\theta^\mu}\mu_i(s|\theta^{\mu})]
\end{split}
\end{equation}
From Eq. \eqref{actor_update}, we can observe that the policy gradient update for the policy ($\mu_s$) of the shared action set $a_s$ is the average of the gradients of the action-value functions of all the tasks it affects. 

This can be easily extended to cases where there are more than one set of shared tasks. Our framework can accommodate heterogeneous dependent action spaces as compared to related multi-task RL algorithms which assume that action spaces are homogeneous or independent or both. This demonstrates the wider applicability of our framework.
\subsection{Algorithm}
\begin{algorithm}[h]
\caption{Multi-task learning using DiGrad}\label{dplr} 
\begin{algorithmic}[1]
\State Randomly initialise actor ($\mu(s|\theta^\mu))$ and critic network ($Q(s,a|\theta^Q$)) with weights $\theta^\mu$ and $\theta^Q$.
\State Initialize the target network with weights \qquad \qquad $\theta^{\mu'}\gets\theta^\mu$ and $\theta^{Q'}\gets\theta^Q$.
\State \textbf{for} $i$ = 1 to E\textsubscript{max} 
\State \quad Initialise random noise $N$ for exploration.
\State \quad Reset the environment to get initial state $s_1$.
\State \quad \textbf{for} $t$ = 1 to Step\textsubscript{max} 
\State \quad \quad Get action $a_t$ = $\mu(s_t|\theta^\mu) + N$.
\State \qquad Execute action $a_t$ and get corresponding reward 
\Statex \qquad vector $\vec{r}_t$ and updated state $s_{t+1}$. 
\State \qquad Store transition ($s_t,a_t,\vec{r}_t,s_{t+1}$) in replay buffer $B$.
\State \qquad Randomly sample a mini-batch $M$ from replay 
\Statex \qquad buffer $B$. 
\State \qquad Update critic $\theta^Q$ according to Eqs.\eqref{loss}, \eqref{target}, \eqref{loss1}.
\Statex \qquad Update actor policy $\theta^\mu$ according to Eq.\eqref{actor_update}
\State \qquad Update the target networks $\theta^{\mu'}$ and $\theta^{Q'}$
\State \quad \textbf{end for}
\State \textbf{end for} 
\end{algorithmic}
\end{algorithm}
In this section we explain the algorithm to learn multiple tasks using DiGrad. The flow of the algorithm is very similar to standard DDPG but there are significant differences in terms of the critic and actor updates as shown in the previous subsections. In DiGrad, compound action $a$ is executed in the environment which returns a vector of rewards $\vec{r}_t$ corresponding to each task instead of a single reward. The replay buffer $B$ stores the current state $s_t$, compound action $a_t$, observed state after executing action $s_{t+1}$ and the vector of rewards $\vec{r}_t$. The entire flow of the algorithm is shown in Algorithm \ref{dplr}. 

\section{Experiments and Results}
The proposed framework was tested in different settings in order to analyse the advantages of each setting. We considered four different network settings for DiGrad as follows:\\ (1) Single critic network with heuristics\\ (2) Single critic network without heuristics\\ (3) Multi critic network with heuristics\\ (4) Multi critic network without heuristics. 

We compare all of them with a standard DDPG setting. We use the same set of hyper parameters  in all the five settings. The critic network architecture is the same for both single and multiple critic case in all aspects except in the number of outputs. The actor network parameters are also same for all the cases. We show the comparison of average reward as well the mean scores of each task in all the plots.  Note that the average reward curves for DDPG are not shown as the reward function settings for DDPG is different than that of DiGrad.

In order to test the proposed multi-task learning framework, we considered two different environments. In both these environments, training involved learning reachability tasks for all the end effectors simultaneously, i.e., learning a policy on the joint space trajectories to reach any point in their workspace.
For all the experiments, we define error and score for a particular task $i$ as,
\begin{equation*}
\begin{split}
error_i = ||G_i - E_i||, \quad score_i = -log(error_i)
\end{split}
\end{equation*}
where $G_i$ and $E_i$ are the coordinates of goal and end-effector of the $i_{th}$ chain.

In all the experiments, agents were implemented using TensorFlow Code base consisting of 2 fully connected layers. RMSProp optimizer is used to train both actor and critic networks with learning rates 0.001 and 0.0001 respectively. We used CReLu activation in all the hidden layers. While training, a normally distributed decaying noise function was used for exploration. Whereas, while testing this noise is omitted. We set the discount factor to be 0.99 in all the settings.  In all the tasks, we use low-dimensional state description which include joint angles, joint positions and goal positions.
The actor output is a set of angular velocities $\dot{q}$. Hence the policy learns a mapping from the configuration space to joint velocity space.
\subsubsection{Reward function}
The reward function for DiGrad settings is modelled keeping in mind the multi-task application. As defined before, $r_i$ is the reward corresponding to the action $a_i$ of the $i_{th}$ task. We give a small positive reward to $r_i$ if task $i$ is finished. Also, if all the end effectors reach their respective goals, a positive reward is given to all the tasks. For all other cases, a negative reward proportional to the $error$ is given. In DDPG setting, there is a single reward unlike DiGrad. A positive reward is given when all the end effectors reach their goals simultaneously. Else, a negative reward is given proportional to the sum of $error$ of all the tasks, that is, sum of distances between the respective goal and its corresponding end effector.
\begin{figure}
\begin{subfigure}{0.25\textwidth}
\includegraphics[width=\columnwidth]{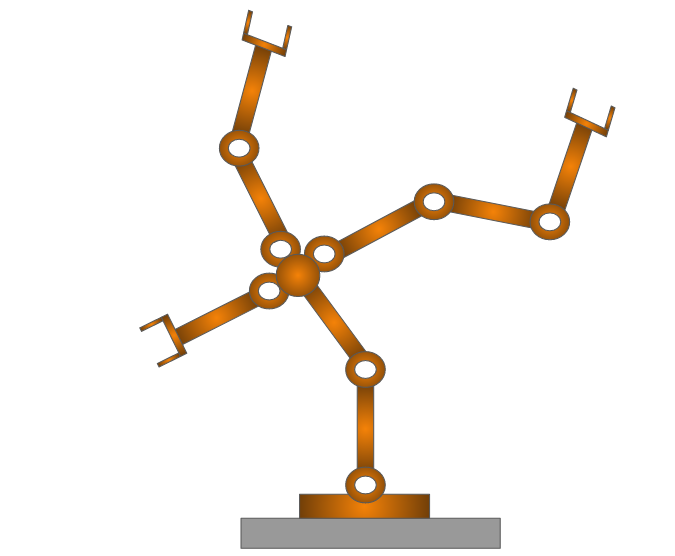}
\caption{8-link Planar Manipulator}
\label{8-link}
\end{subfigure}
\begin{subfigure}{0.2\textwidth}
\includegraphics[width=0.57\columnwidth]{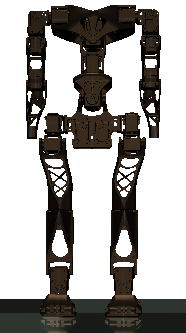}
\caption{Humanoid Robot}
\label{humanoid}
\end{subfigure}
\caption{Environments}
\label{8link}
\end{figure} 
\begin{figure*}[ht]
\includegraphics[width=2.1\columnwidth]{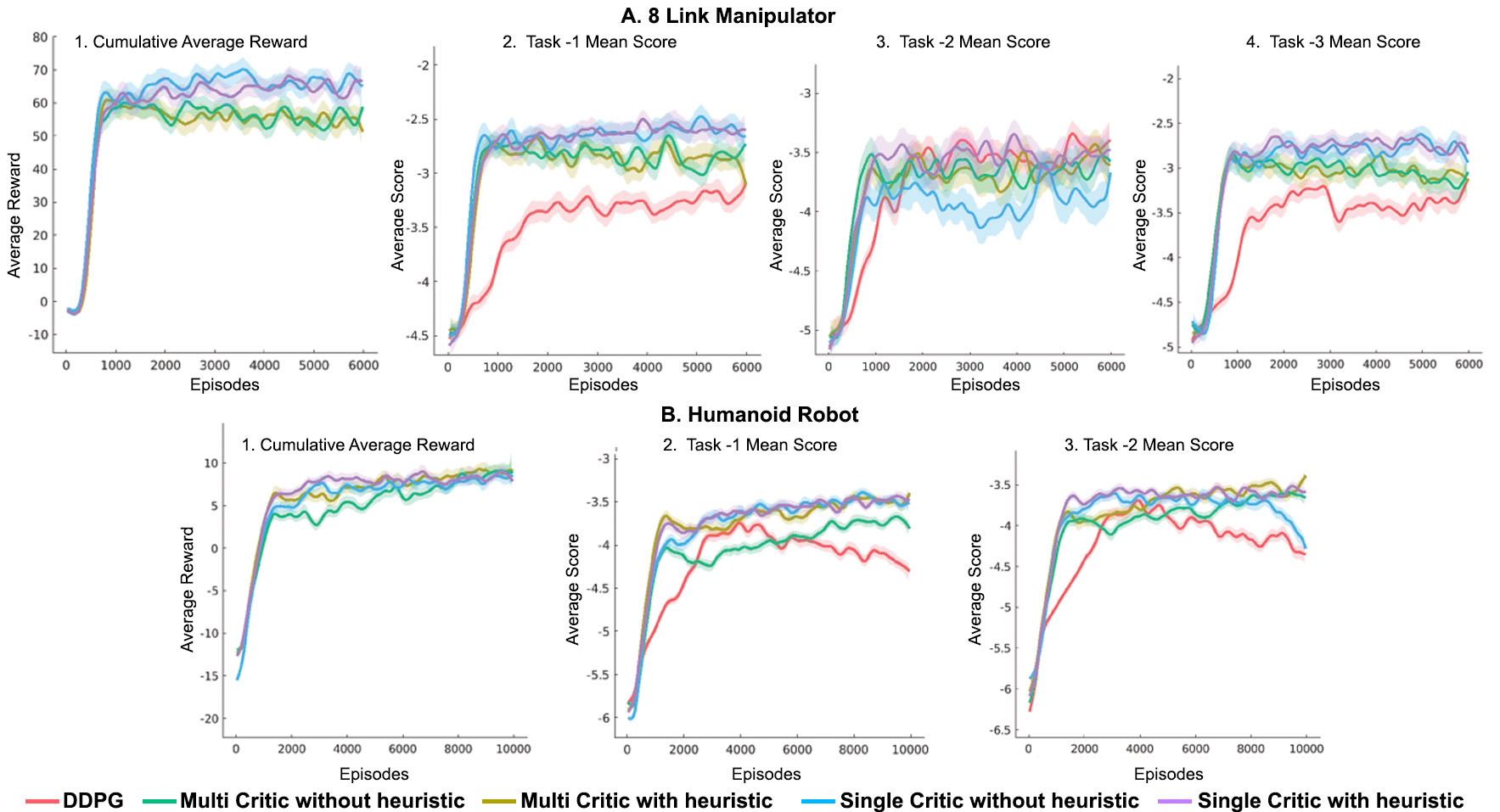}
\caption{Performance curves of reachability task experiments on 8 link manipulator and humanoid. The bold line shows the average over 3 runs and the coloured areas show average standard deviation over the tasks. Note that, average reward curve is not plotted for DDPG as the reward function for it is different from DiGrad frameworks.
}
\label{result_three}
\end{figure*}
\subsubsection{Environments}

\textbf{8-link manipulator}: The first environment is an 8 DoF planar manipulator with 3 end effectors as shown in Fig. \ref{8-link}. We can observe that the shared sub-chain of 2 joints is common to  all the 3 tasks. Also, the action dimension of the non-shared chains are kept different in order to check the robustness of the framework. The dimensions of each action is given as: $a_1 = 3$, $a_2 = 4$, $a_3 = 5$ and $a_s = 2$. 

Fig.\ref{result_three}(A) shows the performance curves for the five different settings. From the mean score curves, we can see that all the DiGrad based settings converge faster and achieve better final performance than DDPG. Even though the action dimension of each task was different, all the network settings in DiGrad framework worked equally well for all the tasks. Whereas, DDPG showed comparable performance for only Task-2. 

From \ref{result_three}(A1), we can see that the single critic framework consistently has better average reward per episode than multi critic frameworks. Thus, modelling all the action value functions in one critic network doesn't affect the performance. In fact, the shared parameters in single critic framework could help the network capture the correlation among the actions much better than multi-critic case. Note that, single critic framework is achieving these performances with lesser parameters than the multi-critic framework. 

DiGrad frameworks with heuristics perform at par with the non-heuristic frameworks. On applying the aforementioned heuristic, no significant improvement in the average reward curve is observed. But in the mean score curves, specially Task-1 and Task-2 curve, we can see that the application of heuristics helps the network to be more stable as compared to their respective non-heuristic curves. Thus, we can say that normalising the gradient of action values of the shared action as in Eq. \eqref{hu} could help the network deliver robust multi-task training.\\
\textbf{Humanoid robot}: Secondly, we test our framework on a 27 DoF humanoid robot (Fig. \ref{humanoid}). This experiment involved reachability tasks of the 2 hands of the humanoid robot using the upper body (13 DoF) consisting of an articulated torso. The articulated torso is the shared chain which is affecting both the tasks. It is noteworthy that the articulated torso has 5 DoF whereas, the arms have 4 DoF each. Thus, the contribution of shared action (articulate torso) to the task is more than the non shared actions (arms). The environment for training is developed in MATLAB and the trained models were tested in a dynamic simulation environment MuJoCo. 

Fig.\ref{result_three}B summarizes the results for this case. We found that DPPG is generally unstable for solving multi-tasks problems. In some runs, it may learn initially but suffers degradation later.  We observe that the DiGrad algorithm yields better final results while having greater stability.
 
From the mean scores of the tasks, we can see that the single critic frameworks converge faster and are stable throughout the experiment as compared to the multi-critic frameworks.  The best setting is again the single critic with heuristic, outperforming all the others in all the cases.\\ \\
Note that, the reward function for DDPG is kept different from the DiGrad framework. We also tried a different reward setting taking the sum of individual rewards $r_i$ as a reward signal to DDPG framework, where $r_i$ is same as defined in the DiGrad reward setting.
We observed that this reward setting led to biasing, where one of the tasks dominated others. This behaviour could have been due to the negative interference across tasks, which didn't happen in DiGrad.
\section{Conclusion}
In this paper, we propose a deep reinforcement learning algorithm called DiGrad for multi-task learning in a single agent. This framework is based on DDPG algorithm and is derived from DPG theorem. In this framework, we have a dedicated action value for each task, whose update depends only on the action and reward corresponding to the task. We introduced a differential policy gradient update for the compound policy.

We tested the proposed framework for learning reachability tasks on two environments, namely 8-link manipulator and humanoid. These experiments show that our framework gave better performance than DDPG, in terms of stability, robustness and accuracy.  These performances are achieved keeping the number of parameters comparable to DDPG framework. Also, the algorithm learns multiple tasks without any decrease in the performance of each task. 

Our work focuses on learning coordinated multi-actions in the field of robotics, where a single agent performs multiple tasks simultaneously. The framework was able to capture the relation among tasks where some tasks had overlapped action space. Our future work will focus on extending the framework to multi-agent domain.  
\newpage
\bibliographystyle{named}
\bibliography{ijcai18}

\end{document}